\documentclass[letterpaper]{article}
  \usepackage{aaai}
  \usepackage{graphicx}
  \usepackage{times}
  \usepackage{amsmath}
  
  \usepackage{helvet}
  \usepackage{courier}
  \usepackage{comment}
  \usepackage{color,soul}
  \usepackage{xcolor}
  \usepackage{todonotes}
  \usepackage[utf8]{inputenc} 
  \usepackage[T1]{fontenc} 
  \usepackage{etoolbox,lineno}
  \usepackage{url}            
  \usepackage{booktabs}       
  \usepackage{amsfonts}       
  \usepackage{nicefrac}       
  \usepackage{microtype}      
  \usepackage{svg}
  \usepackage{subcaption}
  \usepackage{enumitem}
  \usepackage{wrapfig}
  \usepackage{lipsum}
  \usepackage{multirow}
\newcommand{\RowSv}[1]{{\multirow{8}{*}{#1}}}
\usepackage{xcolor}

\begin{document}
  %
  \title{Learning Clusterable Visual Features for Zero-Shot Recognition}
  \author{
Jingyi Xu,\textsuperscript{\rm 1}
Zhixin Shu,\textsuperscript{\rm 2}
Dimitris Samaras\textsuperscript{\rm 1}\\
\textsuperscript{\rm 1}Stony Brook University,
\textsuperscript{\rm 2}Adobe Research\\
Jingyi.Xu.1@stonybrook.edu, zhixinshu@gmail.com, samaras@cs.stonybrook.edu
}
  \maketitle
   \begin{abstract}

     In zero-shot learning (ZSL), conditional generators have been widely used to generate additional training features. These features can then be used to train the classifiers for testing data. However, some testing data are considered ``hard'' as they lie close to the decision boundaries and are prone to misclassification, leading to performance degradation for ZSL. 
     In this paper, we propose to learn clusterable features for ZSL problems. 
     Using a Conditional Variational Autoencoder (CVAE) as the feature generator, we project the original features to a new feature space supervised by an auxiliary classification loss. 
     %
     To further increase clusterability, we fine-tune the features using Gaussian similarity loss.
     The clusterable visual features are not only more suitable for CVAE reconstruction but are also more separable which improves classification accuracy.
     Moreover, we introduce Gaussian noise to enlarge the intra-class variance of the generated features, which helps to improve the classifier's robustness. 
     Our experiments on SUN, CUB, and AWA2 datasets show consistent improvement over previous state-of-the-art ZSL results by a large margin. 
     In addition to its effectiveness on zero-shot classification, we experimentally show that our method to increase feature clusterability benefits few-shot learning algorithms as well.
     
  \end{abstract}
  
\section{Introduction}
  Object recognition has made significant progress in recent years, relying on massive labeled training data.
   However, collecting large numbers of labeled training images is unrealistic in many real-world scenarios.
   For instance, training examples for certain rare species can be challenging to obtain, and annotating ground truth labels for such fine-grained categories also requires expert knowledge.
   Motivated by these challenges, zero-shot learning (ZSL), where no labeled samples are required to recognize a new category, has been proposed to handle this data dependency issue.
    
  Specifically, zero-shot learning aims to learn to classify when only images from seen classes (source domain) are provided while no labeled examples from unseen classes (target domain) are available. 
  The seen and unseen classes are assumed to share the same semantic space, such as the semantic attribute \cite{attribute2009ali,evaluation2015akata,relative2011devi} or word vector space \cite{distributed2013tomas,efficient2013tomas}, to transfer  knowledge between the seen and unseen.
  Existing ZSL learning methods \cite{dap,evalemb2015cvpr,transfer2015cvpr,devise} typically utilize  training data from the source domain to learn a compatible transformation from the visual space to the semantic space of the seen and unseen classes.
  Then for test samples from the target domain, the visual features will be projected into the semantic space using the learned transformation, in which nearest neighbour (NN) search will be conducted to perform zero-shot recognition.
  \begin{figure}
        \begin{subfigure}[b]{0.22\textwidth}
              \includegraphics[width=\linewidth]{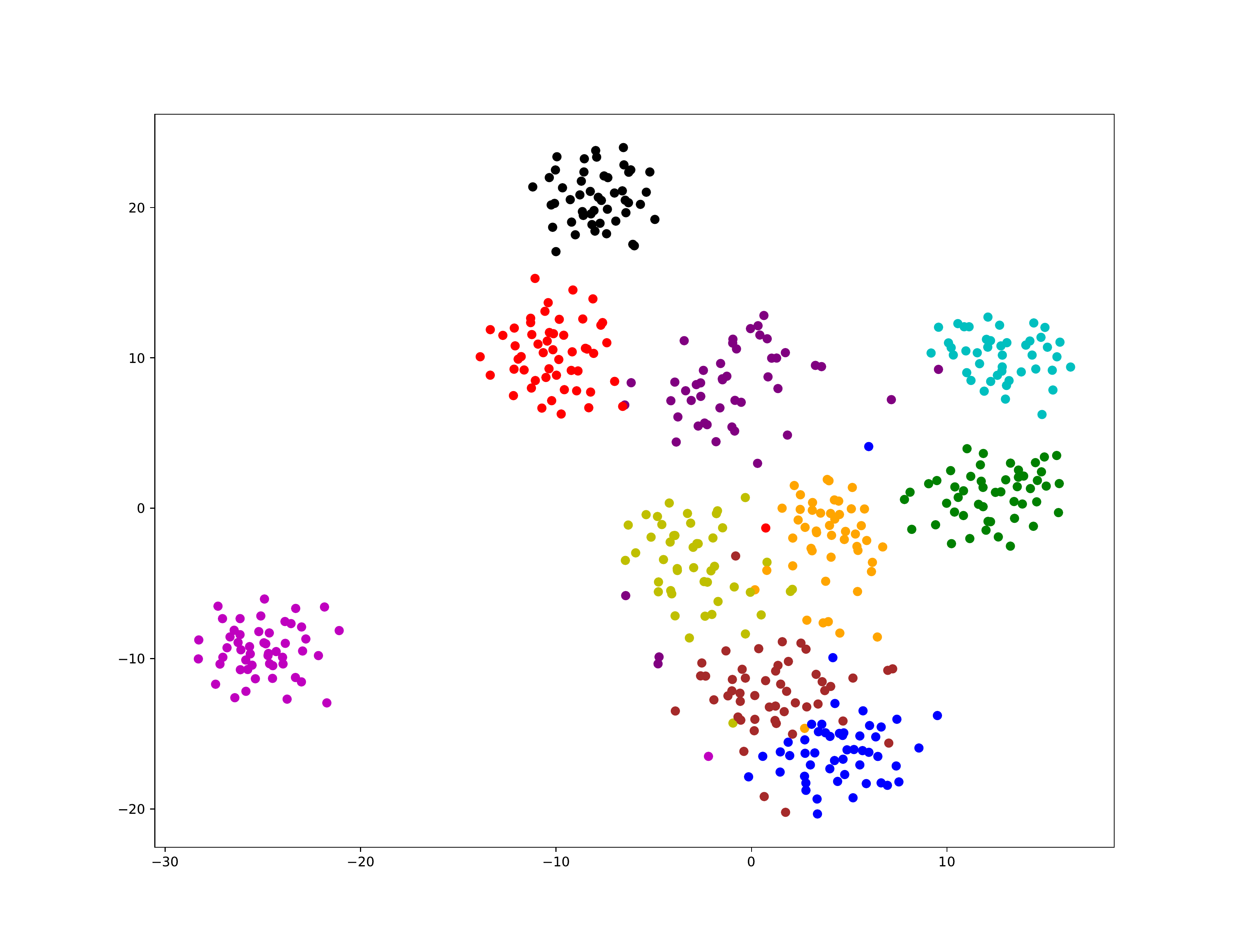}
                \caption{ResNet Features}
                \label{fig:motivation1}
        \end{subfigure}%
        \begin{subfigure}[b]{0.22\textwidth}
           \includegraphics[width=\linewidth]{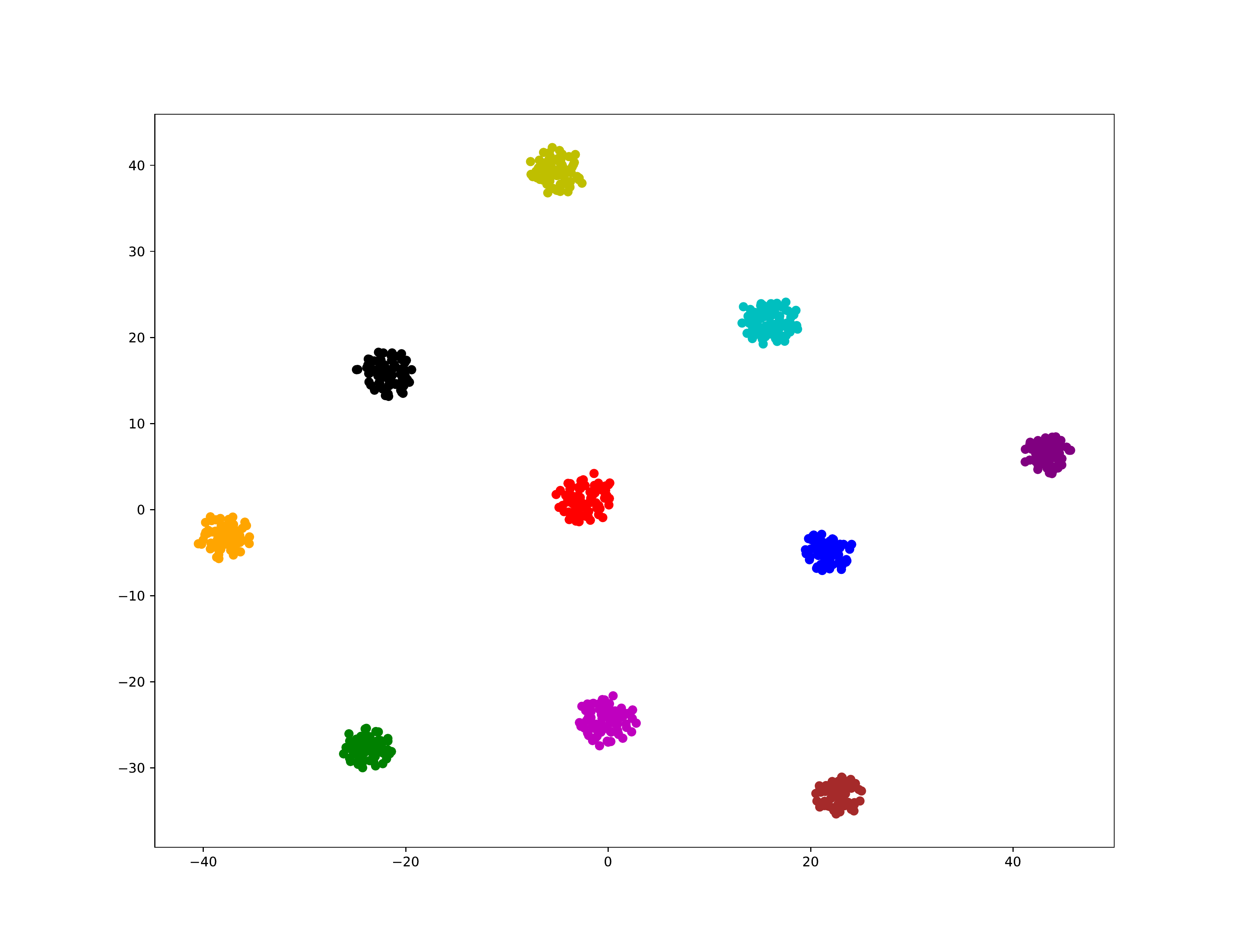}
                \caption{Generated Features}
                \label{fig:motivation2}
        \end{subfigure}%
         \caption{
         Motivation of our method. ResNet features contain hard samples while features synthesized by a CVAE are more clusterable and discriminative.}\label{fig:motivation}
\end{figure}
  In addition to the above setting where the test samples are from target domain only, another more challenging setting is generalized zero-shot learning (GZSL), where the test samples may come from either the source or the target domain.
  For GZSL, 
  %
  a number of methods based on feature generation \cite{cvae-zsl,f-clswgan,se-zsl,cycle2018eccv,f-vaegan-d2} have been proposed to alleviate the data imbalance problem by generating additional training samples with a conditional generator, such as a conditional GAN \cite{f-clswgan,cycle2018eccv,lisgan} or a conditional VAE \cite{se-zsl,cada-vae,cvae-zsl}. 
  A discriminative classifier can be then trained with the artificial features from unseen classes to perform classification.
  %
  %
  
  One of the main limitations of CVAE based feature generation methods is the distribution 
  shift between the real visual features and the synthesized features.  
  The real visual features typically contain hard samples which are close to the decision boundary 
  while the CVAE-generated features are usually more clusterable and discriminative (Figure. \ref{fig:motivation}).
  As a result, the discriminative classifier trained on the generated features can not generalize well on real unseen samples.
  To address the problem, Keshari et al. \cite{ocd} proposed to generate challenging samples that are closer to other competing classes to increase the generalizability of the network.
  %
  %
  In this paper, we try to tackle the problem from a different perspective.
  Instead of generating challenging features to train a robust classifier,
  we propose to project the original visual features 
  to a clusterable and separable feature space.
  The projected features are supervised by a classification loss from a discriminative classifier.
  %
  The advantages of learning clusterable visual features are two-fold: first, since the features generated by CVAE are clusterable in nature \cite{cvae-zsl}, the projected features will be more suitable for CVAE reconstruction;
  and second, during testing, the test samples are easier to be classified correctly after being projected to the same discriminative feature space using the learned mapping function compared to the original hard ones.
  
  To further increase the visual features' clusterability, we utilize Gaussian similarity loss, which was first proposed by Kenyon-Dean \textit{et al.} \cite{corel}, to fine-tune the visual features before the CVAE reconstruction.
  The fine-tuning step helps to derive a more clusterable feature space and further improves ZSL performance.
 
  %
  
  In addition to learning clusterable visual features,
   we generate hard features to improve the robustness of the classifier.
   In practice, we introduce Gaussian noise when minimizing reconstruction loss to synthesize features with larger intra-class variance.
   In experiments, we show that our method, which simultaneously increasing the feature's clusterability and the classifier's generalizability, consistently improves the state-of-the-art zero-shot learning performance on various datasets significantly.
   Remarkably, we achieve 16.6\% improvement on SUN dataset and 12.2\% improvement on AWA2 dataset over previous best accuracy of unseen classes in the GZSL setting.
   In addition to zero-shot learning  scenarios, we also apply our method on few-shot learning problems and show that more clusterable features can benefit few-shot learning as well.
   

  In summary, our contributions are as follows:
  \begin{itemize}
    \item For CVAE based feature generating methods, we propose to increase the clusterability of real features by projecting the original features to a discriminative feature space.
    The projected features can better mimic the distribution of generated features and are also easier to classify, which leads to better zero-shot learning performance.
    \item We utilize Gaussian similarity loss to increase the clusterability of visual features and experimentally demonstrate that more clusterable features benefit both zero-shot and few-shot learning.
    \item On both ZSL and GZSL settings, our method significantly improves the state-of-the-art zero-shot classification performance on CUB, SUN and AWA2 benchmarks.
  \end{itemize}
 \begin{figure*}
  \centering
  \includegraphics[width=140mm]{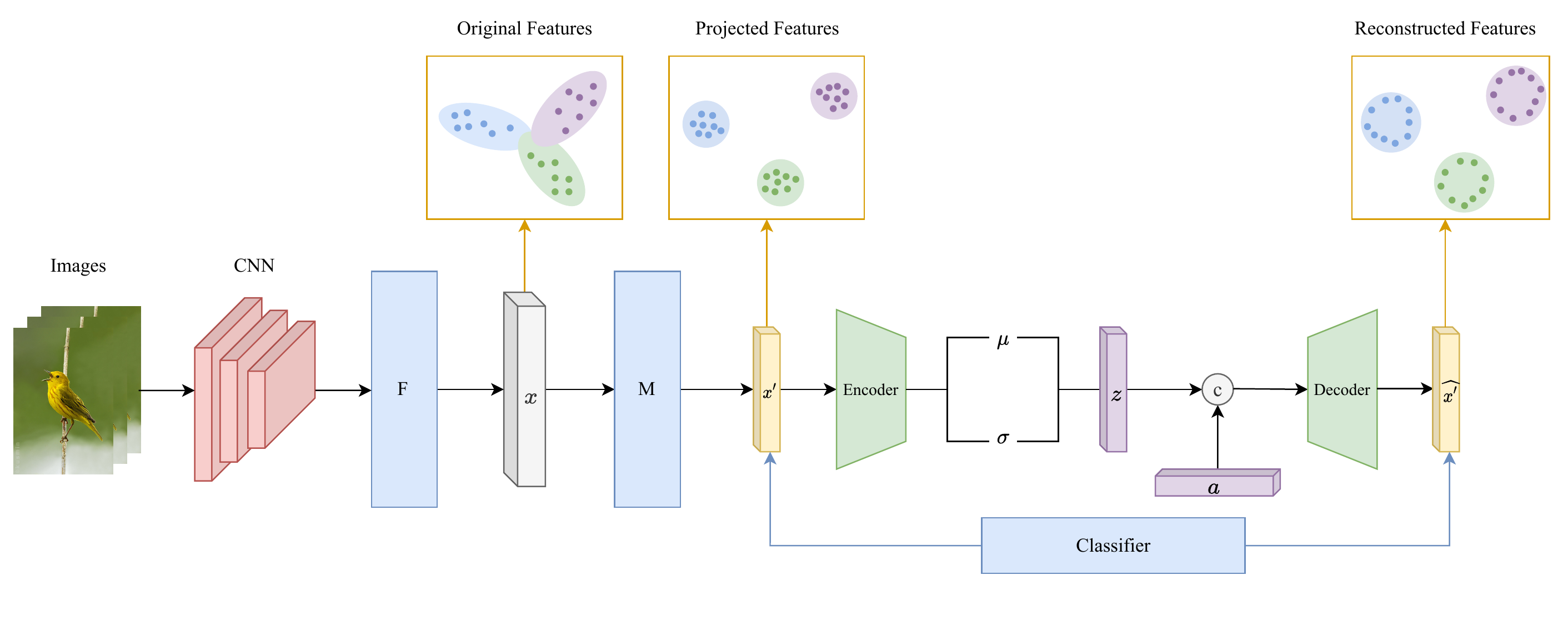}\\
   \caption{\textbf{The pipeline of our proposed method}. Instead of reconstructing the original visual features directly, we use a mapping function $M$ to project the original visual features to a clusterable feature space. 
   Another mapping function $F$ is introduced for fine-tuning. 
   The projected features are supervised by the classification loss produced by a classifier and the reconstructed features are supervised by the same classifier to enforce the same distribution.
   The synthesized features are expected to exhibit larger intra-class variance, which will lead to a more robust classifier.
  Best viewed in color.
  }
  \label{fig:env}
\end{figure*}
 \section{Related Work}
 \subsection{Clustering-based Representation Learning}
 Clustering-based representation learning has attracted interest recently.
 A discriminative feature space is desirable for many image classification problems, especially   
 facial recognition.
 Schrof \textit{et al.} proposed a triplet loss \cite{triplet-loss2015}, which minimizes the distance between an anchor and a positive sample while maximizing the distance between an anchor and a negative sample until a margin is met.
 However, since image triplets are the input of the network, carefully designed methods to sample training data are required.
 Another line of research avoids such a pair selection procedure by using a softmax classifier, with a variety of loss functions designed to enhance discriminative power \cite{arcface2019,center-loss,cosface2018,sphereface}.
 Wen \textit{et al.} proposed a center loss \cite{center-loss}, which penalizes the Euclidean distance between each feature vector and its class center to increase intra-class compactness.
 %
 %
 Besides the domain of facial recognition,  Kenyon-Dean \textit{et al.} proposed clustering-oriented representation learning (COREL) 
 which builds latent representations that exhibit the quality of natural clustering.
 The more clusterable latent space leads to better classification accuracy for image and news article classification.
 In this paper, we empirically show that more clusterable visual features can benefit zero-shot learning and few-shot learning as well. 
 %
 \subsection{ZSL and GZSL}
 Conventional zero-shot learning methods generally focus on learning robust visual-semantic embeddings\cite{eszsl,devise,metric2016bucher,sae}.
 Lampert \textit{et al.}\cite{dap} proposed Direct Attribute
 Prediction (DAP), in which a probabilistic classifier is learned for each attribute independently.
 The trained estimators can then be used to map attributes to the class label at the inference stage.
 Bucher \textit{et al.} \cite{metric2016bucher} proposed to control the semantic embedding of images
 by optimizing jointly the attribute embedding and the classification metric in a multi-objective framework.
 Kodirov \textit{et al.}  proposed a Semantic Autoencoder (SAE) \cite{sae}, which uses an additional reconstruction constraint to enhance ZSL performance.
 
 In the more challenging GZSL task, in which test samples can be from either seen or unseen categories, semantic embedding methods suffer from extreme data imbalance problems.
 The mapping between visual and semantic spaces has a bias towards the semantic features of seen classes, thus hurting the classification of unseen classes significantly.
 Recent research address the  lack of training data for unseen classes 
 by synthesizing visual representations via generative models \cite{f-clswgan,cvae-zsl,lisgan,se-zsl,ocd,generating_representation_zsl,dual_gzsl,rethinking2019kai}.
 Xian \textit{et al.} \cite{f-clswgan} propose to generate image features with a WGAN conditioned on class-level semantic information. %
 The generator is coupled with a classification loss to generate sufficiently discriminative CNN features.
 %
 %
 LisGAN\cite{lisgan}, built on top of WGAN, employs
 `soul samples' as the representations of each category to improve the quality of generated features. 
 
  However, GAN-based losses suffer from mode collapse issues and instability in training.
  Hence, conditional variational autoencoders (CVAE) have been employed for stable training\cite{se-zsl,zsl_generating_learning,cvae-zsl}. 
  Verma \textit{et al.} \cite{se-zsl} incorporates a CVAE based architecture with a discriminator that learns a mapping from the VAE generator's output to the class-attribute, leading to an improved generator. 
  Yu \textit{et al.} \cite{zsl_generating_learning} leverages a CVAE with category-specific multi-modal prior by generating and learning simultaneously.
  The trained CVAE is provided with experience about both seen and unseen classes. 
  We also deploy a CVAE to synthesize features for the unseen classes. Our method’s novelty lies in that we project the original visual features to a clusterable and discriminative feature space. The projected features are more suitable for CVAE reconstruction and easier for the final classifier to classify correctly.

 \section{Method}
  
  We first generate additional training samples with a CVAE-based architecture as the baseline of our model.
  In the following we describe how we learn clusterable visual features by using a discriminative classifier as well as fine-tuning with Gaussian similarity loss.
  Finally, we describe how we obtain a more robust classifier by synthesizing hard features to further improve ZSL performance.
  
  \textbf{Problem Setup:} In zero-shot learning, we have a training set $S_{tr}={(x_i, y_i, a_i)}$ where $x_i \in X^S$ are the visual features, $y_i \in Y^S = \{y_1, ... y_K\}$ denotes the class labels of source seen classes and $a_i$ is the semantic descriptor, e.g., semantic attributes, of class $y_i$.
  In addition, we have a set of target unseen class labels $Y^U = \{u_1,..., u_L\}$ which have no overlap with the source seen classes, \textit{i.e.}$Y^S \cap Y^U = \emptyset$.
  For unseen classes, we are given their semantic features but their visual features are missing.
  Zero-shot learning methods aim to learn a model which can classify the datapoints
  from unseen classes $x_i \in X^U$ labeled $y_i \in Y^U$.
  \subsection{Baseline ZSL with Conditional Autoencoder}
  The conditional VAE proposed in \cite{cvae-zsl} consists of a probabilistic encoder model $E$ and a probabilistic decoder model $G$.
  The encoder $E$ takes the input sample $x_i$ as input and encodes the latent variable $z_i$.
  The encoded variable $z_i$, concatenated with the corresponding attribute vector $a_i$, is provided to the decoder $G$, which is trained to reconstruct the input sample $x$.
  The training loss is given by: 
  \begin{equation}\label{eq:cvae}
  \begin{aligned}
      L_{CVAE} = &-E_{p_{E(z|x)}, p(a|x)}[\textnormal{log}p_G(x|z,a)] + \\ &\textnormal{KL}(p_E(z|x)||p(z))
  \end{aligned}
  \end{equation}
  The first term is the generator's reconstruction loss and the second term is the KL divergence loss that pushes the VAE posterior to be close to the prior.
  The latent code $z_i$ represents the class-independent component and the attribute vector $a_i$ represents the class-specific component.
  %
  
  Once the VAE is trained, one can synthesize samples of any class by sampling from the prior $p(z)$, specifying the class attribute vector $a$ and generating samples $\widehat{x}$ with the generator.
  The generated samples can then be used to train a discriminative classifier, such as a support vector machine (SVM) or a softmax classifier. 
  %
   
  \subsection{Discriminative Embedding Space for Reconstruction}
  Instead of training a CVAE to mimic the distribution of real visual features directly,
  we project the original features to a clusterable and discriminative feature space and try to reconstruct the projected features instead.
  The projected features are expected to have low intra-class variance and large inter-class distance,  mimicking the CVAE feature distribution.
  To ensure such a distribution, we introduce a discriminative classifier and minimize the classification loss over the projected features.
  
  \begin{equation}\label{eq:cls_loss}
   L_{cls} = - E_{p_M(x'|x)}[\textnormal{log}p_C(y|x')],
  \end{equation}
  
  \noindent where $M$ is the mapping function and $x'$ denotes the projected features.
  $C$ is the discriminative classifier and $p_C(y|x')$ is the probability of predicting $x'$
 with its true label $y$.

    To further enforce the same distribution between the projected features and the reconstructed features, we minimize the classification error of the same classifier over the reconstructed features as well:
    \begin{equation}
  \begin{aligned}
   L_{cls'} = - E_{p_G(\widehat{x'}|z, a)}[\textnormal{log}p_C(y|\widehat{x'})],
  \end{aligned}
\end{equation}
  The complete learning objective is given by:
  \begin{equation}\label{eq:obj}
  \begin{aligned}
      \min_{\theta_E, \theta_G, \theta_M, \theta_C} L_{CVAE} + \lambda_{cls}*L_{cls} + \lambda_{cls'}*L_{cls'},
  \end{aligned}
  \end{equation}
  
 \noindent where $\lambda_{cls}$ and $\lambda_{cls'}$ are the loss weights for the classification losses produced by the projected features and the reconstructed features respectively.
  
  Finally in the testing stage, the test samples will also be projected to the discriminative space using the same mapping function $M$ to be classified by the discrminative classifier.
  Compared to the original samples, the projected samples are easily separated and more likely to be classified correctly.
  
  \subsection{Clusterable Feature Learning with Gaussian Similarity Loss}
  The Gaussian similarity loss was first proposed in \cite{corel} to create representations which exhibit the quality of natural clustering.
  Here we adopt the Gaussian similarity loss to fine-tune the visual features to further increase the clusterability of the feature space.

  Neural networks for conventional classification tasks are trained using a categorical cross-entropy (CCE) loss. 
  Specifically, the CCE loss seeks to maximize the log-likelihood of
  the $N$-sample training set from $K$ classes:
  \begin{equation}
  \begin{aligned}
      L_{CCE} &= -\sum_{i=1}^{N} \textnormal{log} \frac{\textnormal{exp}(w_{y_i}^T h^i)}{\sum_{k=1}^{K}\textnormal{exp}(w_k^Th^i)}, \\
  \end{aligned}
  \end{equation}
  where $h^i$ is the $i$th feature sample with label $y_i$ and $w_k$ is the $k$th column of the classification matrix $W$.
   We algebraically reformulate the CCE loss formulation as follows:
   \begin{equation}
  \begin{aligned}
      L_{CCE} &= \sum_{i=1}^N -s(h^i, w_{y_i}) + \textnormal{log} \sum_{k=1}^K
      e^{s(h^i, w_k)},
  \end{aligned}
  \end{equation}
  where $s(h^i, w_{y_i}) = w_{y_i}^T \cdot h^i$ is the similarity function between $h^i$ and $w_{y_i}$, which is the dot product in the CCE loss.
  
  Although the CCE loss is widely used for classification tasks,
  the representations learned by CCE are not naturally clusterable.
  Replacing the dot product operation in the CCE loss with Gaussian similarity function,
  we get the Gaussian similarity loss which leads to more clusterable latent representations \cite{corel}.
  Specifically, the Gaussian similarity function is defined based on the univariate normal probability density function and the standard radial basis function (RBF) as follows:
  
   \begin{equation}
   \begin{aligned}
     s(h^i, w_{y_i}) = - \gamma \| h^i - w_{y_i} \|^2,
  \end{aligned}
  \end{equation}
  
  \noindent where the hyper parameter $\gamma$ is a free parameter.
  Thus, the Gaussian similarity loss can be written as:
  \begin{equation}
   \begin{aligned}
      L_{GAU} &= \sum_{i=1}^N -s(h^i, w_{y_i}) + \textnormal{log} \sum_{k=1}^K e^{s(h^i, w_k)} \\
      &= \sum_{i=1}^N \gamma \| h^i - w_{y_i} \|^2 + \textnormal{log} \sum_{k=1}^K e^{-\gamma\|h^i - w_{y_i}\|^2}
  \end{aligned}
  \end{equation}
  
  According to \cite{corel}, compared to the  CCE loss, the Gaussian similarity loss helps to create naturally clusterable latent spaces.
  To fine-tune the original visual features with the Gaussian similarity loss, we use another mapping function $F$ to transform the original features $x$ to a new space and minimize the Gaussian similarity loss of the transformed features $F(x)$ w.r.t a new classification matrix $W'$: \begin{equation}
  \begin{aligned}
      \min_{\theta_F, \theta_W'} L_{GAU}(F(x), W')
  \end{aligned}
  \end{equation}
  The transformed features $F(x)$ can be then used for CVAE reconstruction.
 \begin{table*}[t] 
    
    \centering
  \resizebox{0.75\textwidth}{!}{%
    \begin{tabular}{l|l|ccc|ccc|ccc}
      \hline
      & Method & \multicolumn{3}{c|}{CUB} & \multicolumn{3}{c|}{SUN} &
                \multicolumn{3}{c}{AWA2}  \\
      & & U & S & H & U & S & H & U & S & H\\
      \hline
      \RowSv{$\dag$}& SJE \cite{sje}    & 23.5  & 59.2  & 33.6  & 14.7  & 30.5 & 19.8 & 8.0  & 73.9 & 14.4\\
      & ESZSL \cite{eszsl}        & 12.6 & 63.8 & 21.0  & 11.0 & 27.9 & 15.8 & 5.9 & 77.8 & 11.0\\
      & ALE \cite{ale}            & 23.7 & 62.8 & 34.4   & 21.8 & 33.1  & 26.3 & 14.0 & 81.8 & 23.9\\ 
      & SAE \cite{sae}       & 7.8 & 54.0 & 13.6  & 8.8  & 18.0  & 11.8 & 1.1 & 82.2 & 2.2\\
      & SYNC \cite{sync}                           & 11.5& 70.9 & 19.8  & 7.9 & 43.3  &13.4 & 10.0 & 90.5  & 18.0 \\
      & LATEM \cite{latem} & 15.2 & 57.3 & 24.0  &  14.7 & 28.8  & 19.5  & 11.5 & 77.3 & 20.0\\
      & DEM \cite{dem} & 19.6 & 57.9  & 29.2  &  20.5 & 34.3  & 25.6 &  30.5 & 86.4  & 45.1 \\
      & AREN \cite{aren} & 38.9 & 78.7  & 52.1 &  19.0 & 38.8  & 25.5 &  15.6  & 92.9 & 26.7 \\
      & DEVISE \cite{devise}  & 23.8  & 53.0 & 32.8 & 16.9 & 27.4 & 20.9 & 17.1 & 74.7 & 27.8 \\ \hline
      \RowSv{$\ddag$}& SE-ZSL \cite{se-zsl}   & 41.5  & 53.3  & 46.6 & 40.9 & 30.5  & 34.9 & 58.3 & 68.1 &  62.8 \\
      & f-CLSWGAN \cite{f-clswgan}   & 41.5  & 53.3  & 46.6 & 40.9 & 30.5  & 34.9 & 58.3 & 68.1 &  62.8 \\
      & CADA-VAE \cite{cada-vae}   & 51.6  & 53.5  & 52.4 & 47.2 & 35.7  & 40.6 & 55.8 & 75.0 &  63.9 \\
      & JGM-ZSL \cite{jgm-zsl}  & 42.7  & 45.6  & 44.1 & 44.4 & 30.9  & 36.5 & 56.2 & 71.7 &  63.0 \\
      & RFF-GZSL \cite{rff-gzsl}  & 52.6  & 56.6  & 54.6 & 45.7 & 38.6  & 41.9 & - & - &  - \\
       & LisGAN \cite{lisgan}  & 46.5  & 57.9  & 51.6 & 42.9 & 37.8  & 40.2 & 47.0 & 77.6 &  58.5 \\
      & OCD \cite{ocd}  & 44.8  & 59.9  & 51.3 & 44.8 & 42.9  & 43.8 & 59.5 & 73.4 &  65.7 \\
      & Ours   & \textbf{56.8} & \textbf{69.2}& \textbf{62.4}  &  \textbf{63.8} & \textbf{45.4}  &  \textbf{53.0}  & \textbf{71.6} & \textbf{87.8}& \textbf{78.8}\\
      \hline
    \end{tabular}}  \\ \vspace{6pt}
    \caption{Generalized zero-shot learning performance on CUB, SUN and AWA2 dataset.
    U = Top-1 accuracy of the test unseen-class samples, S = Top-1 accuracy of the test seen-class samples, H = harmonic mean.
    We measure top-1 accuracy in \%. The best performance is indicated in bold.
    }\label{tab:gzsl}%
    \vspace{-10pt}
  \end{table*}
  \begin{table}[t]
  \scriptsize
    \centering
    \renewcommand{\tabcolsep}{2mm}
    \begin{tabular}{l|c|c|c}
      \hline
      Method & {CUB} & {SUN} & {AWA2}  \\
      \hline
      SSE \cite{sse}    & 43.9  & 51.5 & 61.0 \\
      ALE \cite{ale}   & 54.9 & 58.1  & 62.5 \\
      DEVISE  \cite{devise}       & 52.0  & 56.5  & 59.7 \\
      SJE  \cite{sje}  & 53.9    & 53.7  & 61.9\\ 
      ESZSL \cite{eszsl} & 53.9   & 54.5  & 58.6 \\
      SYNC \cite{sync} & 55.6   & 56.3  & 46.6\\ 
      SAE \cite{sae}  & 33.3   & 40.3 & 54.1 \\
      GFZSL \cite{gfzsl}  & 49.2  & 62.6 & 67.0\\
      SE-ZSL \cite{se-zsl}  & 59.6  & 63.4 & 69.2\\
      LAD  \cite{lad}  & 57.9& 62.6 & 67.8\\
      CVAE-ZSL \cite{cvae-zsl} & 52.1 & 61.7  & 65.8\\
      CDL \cite{proto2018eccv}     & 54.5  & 63.6 & 67.9\\
      OCD  \cite{ocd} & 60.3 & 63.5 & 71.3\\\hline
      Ours  & \textbf{63.1}  &  \textbf{65.5} &  \textbf{74.1}\\
      \hline
    \end{tabular}  \\ \vspace{6pt}
    \caption{Classification accuracy for conventional zero-shot learning for the proposed split (PS) on CUB, SUN and AWA2. 
    The best performance is indicated in bold.
    }\label{tab:conventional_zsl}%
    \vspace{-10pt}
  \end{table}
   \subsection{Introducing Gaussian Noise for Better Generalizability}
  
  With the above mapping function supervised with classification loss and fine-tuning with Gaussian
  similarity loss, we obtain discriminative and clusterable real visual features.
 Meanwhile, we introduce Gaussian noise when minimizing the reconstruction loss to synthesize hard features:
  \begin{equation}\label{eq:noise}
    \begin{aligned}
        \widetilde{x'} = x' + \alpha *\epsilon, \epsilon \sim N(0,1)
      \end{aligned} 
  \end{equation}

  \noindent where $x'$ is the projected features and $\widetilde{x'}$ are the features permuted with Gaussian noise; $\alpha$ denotes the strength of the permutation.
  Instead of reconstructing $x'$, the decoder is trained to reconstruct $\widetilde{x'}$.
  The synthesized features will have larger intra-class variance and thus lead to a more rebust final classifier.
  The CVAE training loss will then become:
  \begin{equation}
  \begin{aligned}
    L_{CVAE} = &-E_{p_{E(z|x)}, p(a|x)}[\textnormal{log}p_G(\widetilde{x'}|z,a)] + \\ &\textnormal{KL}(p_E(z|x)||p(z))
  \end{aligned}
  \end{equation}

  \section{Experiments}
  
  We compare our proposed framework in both ZSL and GZSL settings on three 
  benchmarking datasets: CUB \cite{cub}, SUN \cite{sun2012cvpr} and AWA2 \cite{dap}.
  We present datasets, evaluation metrics, experimental
  results and comparisons with the state-of-the-art.
  Moreover, we perform few-shot learning experiments on CUB and SUN dataset 
  to further demonstrate that the clusterable features fine-tuned with Gaussian similarity
  loss can benefit few-shot learning as well.

  \subsection{ZSL and GZSL}
  \subsubsection{Datasets:}
  Caltech-UCSD-Birds 200-2011 (CUB) \cite{cub} and SUN Attribute (SUN)
  are both fine-grained datasets.
  CUB contains 11,788 examples of 200 fine-grained species annotated with 312 attributes.
  SUN consists of 14,340 examples of 717 different scenes annotated with 102 attributes.
  We use a split of 150/50 for CUB and 645/72 for SUN respectively.
  The Animals with Attributes2 (AWA2) \cite{dap} is a coarse-grained dataset proposed for animal classification.
  It is the extension of the AWA \cite{awa2009cvpr} database containing 37,322 samples from 50 classes and 85
  attributes.
  We adopt the standard 40/10 zero-shot split in our experiments.
  The statistics and protocols of the datasets are presented in Table \ref{tab:datasets}.

  \subsubsection{Evaluation Protocol:}

  We report results on the  split 
  (PS) proposed by Xian \textit{et al}. \cite{good_bad_ugly}, which guarantees that no target classes are from ImageNet-1K since it is used to pre-train the base network.
  For ZSL, we adopt the same metrics, i.e., top-1 per class accuracy, as \cite{good_bad_ugly} for fair comparison with other methods.
  %
  %
  For GZSL, we compute top-1 per class accuracy on seen classes, denoted as $S$,
  top-1 per class accuracy on unseen classes, denoted as $U$, and their harmonic mean,
  defined as $H = (2 * S * U) / (S + U)$.
  
  \begin{table}[t]
    \centering
    \renewcommand{\tabcolsep}{1mm}
    \begin{tabular}{l|c|c|c}
      \hline
      Dataset & {Attribute-Dim} & {Images} & {Seen/Unseen Classes}  \\
      \hline
      CUB     & 312  & 11788 & 150/50 \\
      SUN  & 102 & 14340  & 645/72 \\
      AWA2  & 85  &  37322 & 40/10 \\
      \hline
    \end{tabular}  \\ \vspace{6pt}
    \caption{Datasets used in our experiments
    }\label{tab:datasets}%
    \vspace{-10pt}
  \end{table}

\begin{figure*}
        \begin{subfigure}[b]{0.33\textwidth}
                \includegraphics[width=1.1\linewidth]{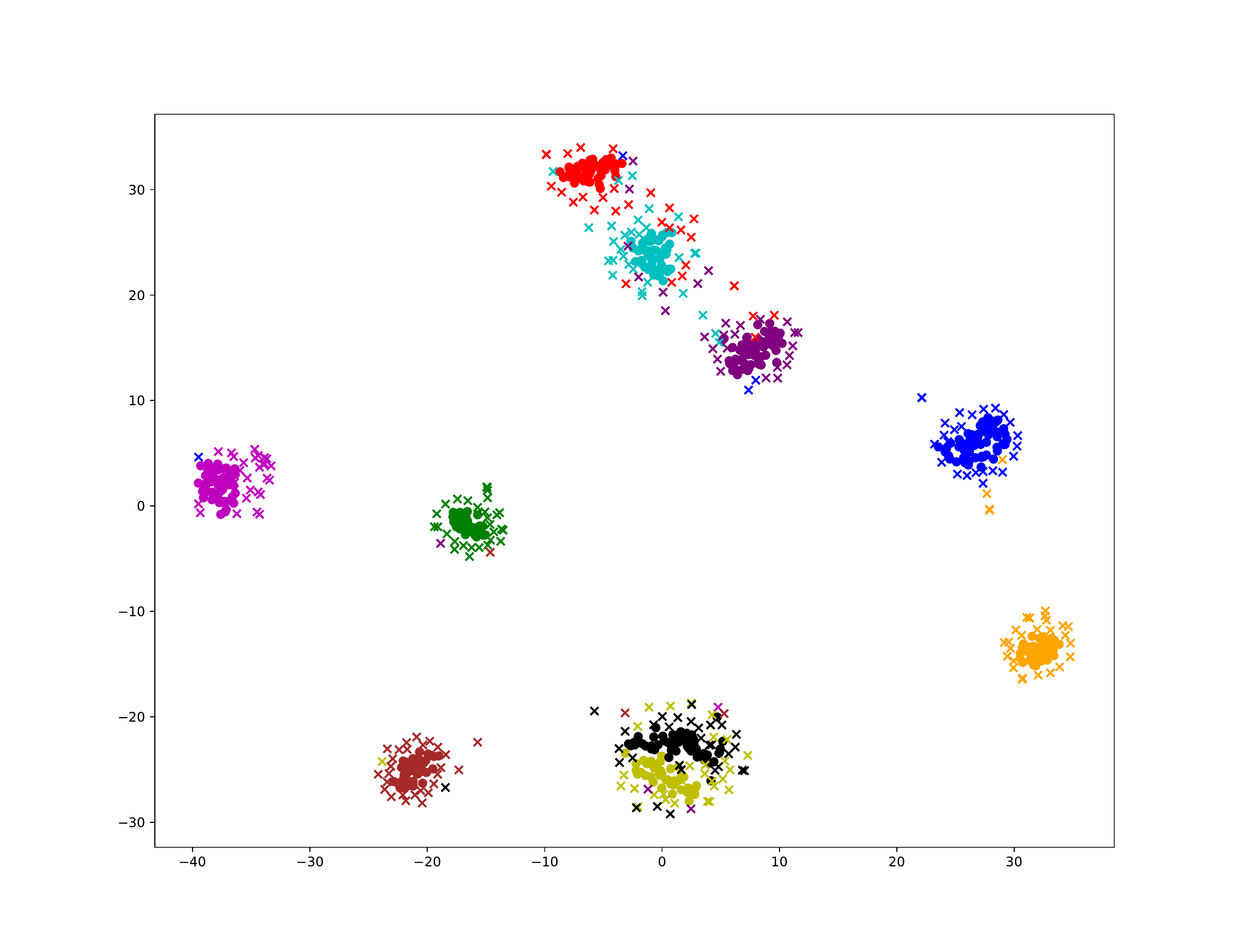}
                \caption{}
                \label{fig:vis1}
        \end{subfigure}%
        \hspace{-3pt}
        \begin{subfigure}[b]{0.33\textwidth}
                \includegraphics[width=1.1\linewidth]{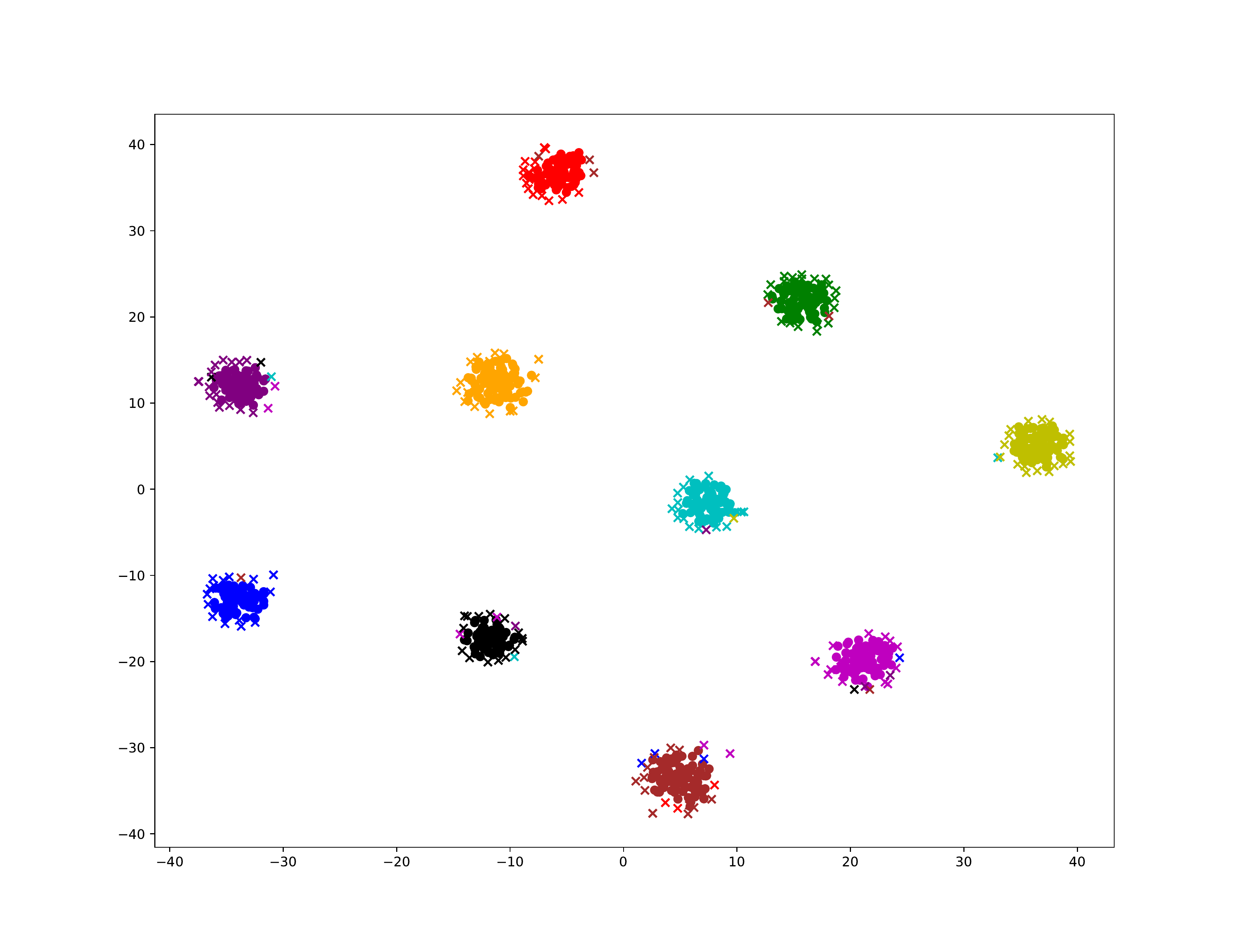}
                \caption{}
                \label{fig:vis2}
        \end{subfigure}%
        \hspace{-3pt}
        \begin{subfigure}[b]{0.33\textwidth}
                \includegraphics[width=1.1\linewidth]{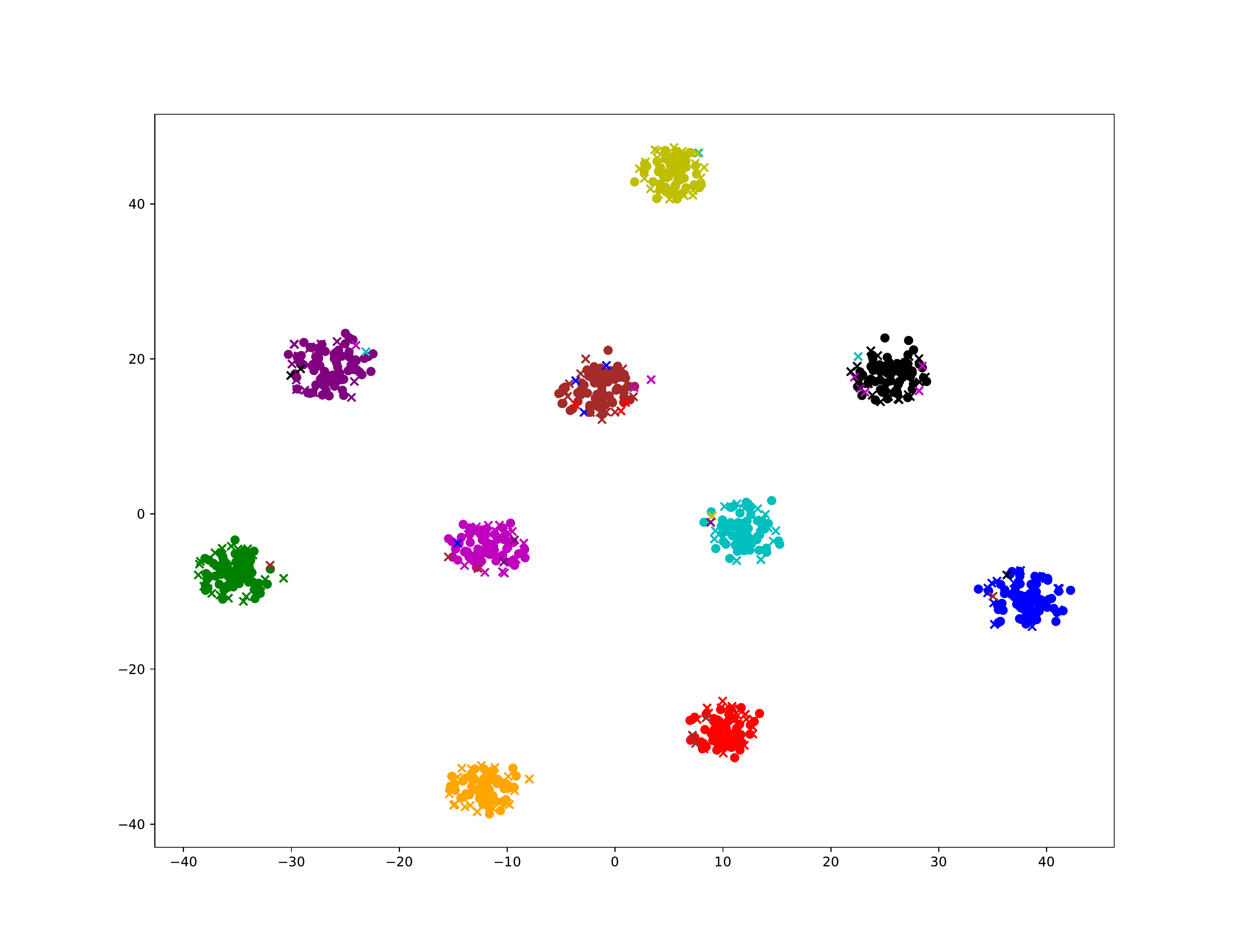}
                \caption{}
                \label{fig:vis3}
        \end{subfigure}%
         \caption{Visualization of real visual features and synthesized features on CUB dataset using t-SNE. (a) Original real features and synthesized features. (b) Projected real features and synthesized features. (c) Projected real features and synthesized features via introducing Gaussian noise. The real features are represented by `$\times$` and the synthsized features are represented by '$\bullet$'.
  Different colors represent different categories. With our proposed method, the real visual features are more discriminative and easily separated while the synthesized features exhibit larger intra-class variance.
  }\label{fig:visualization}
\end{figure*}

  \subsubsection{Implementation Details:}
  Our method is implemented in PyTorch.
  We extract  image features using the ResNet101 model \cite{resnet} pretrained
  on ImageNet with  224 * 224 input size.  
  The extracted features are from the 2048-dimensional final pooling layer.
  Both the encoder $E$ and the decoder $G$ are multilayer perceptrons with one 4096-unit hidden layer.
  LeakyReLU and ReLU are the nonlinear activation functions in the hidden and  output
  layers respectively. 
  The mapping function $M$ is implemented with a fully connected layer and Sigmoid activation.
  The dimension of the latent space is chosen to be the same as the attribute vector.
  For CUB and SUN dataset, the dimension of the projected visual feature space is set to be 512 while for AWA2 dataset, it is set to be 2048.
  We use the Adam solver with $\beta_1 = 0.9$, $\beta_2 = 0.999$ and a learning rate of 0.001.
  We set $\lambda_{cls} = 1$, $\lambda_{cls'} = 0.1$ in eq. \ref{eq:obj}.
  For the CUB and SUN datasets, we set $\alpha = 0.2$ in eq.\ref{eq:noise}.
  For the AWA2 dataset, $\alpha$ is set to 1.
  \subsubsection{Conventional Zero-Shot Learning(ZSL):}

  Table \ref{tab:conventional_zsl} summarizes the results of conventional Zero-Shot Learning.
  %
  %
  In these experiments,  test samples can only belong to the unseen categories $Y^U$.
  It can be seen that our proposed method achieves state-of-the-art performance on all three datasets.
  The classification accuracies obtained on the
  PS protocol on CUB, SUN and AWA2 are 63.1\%, 65.5\%, and 74.1\%, respectively.
  The proposed method improves the state-of-the-art performance by 2.9\%
  on CUB, by 2.0\% on SUN and by 2.8\% on AWA2, which indicates the effectiveness of the framework.
  Our performance beats other CVAE based zero-shot learning methods, such as SE-ZSL\cite{se-zsl} and CVAE-ZSL \cite{cvae-zsl}, by a large margin.
  Compared to \cite{ocd} which synthesizes hard features to increase network generalizability , we increase the clusterability of real visual features at the same time to make it easily separated and more reconstructable, which leads to a significant accuracy boost.
  
  \subsubsection{Generailized Zero-Shot Learning(GZSL):}
   In the GZSL setting, the testing samples can be from either seen or unseen classes.
  The setting is more challenging and more reflective of real-world application scenarios, since typically whether an image is from a source or target class is unknown in advance.

  Table \ref{tab:gzsl} shows the generalized zero-shot recognition results on the three datasets.
  We group the algorithms into non-generative models, i.e., SJE \cite{sje}, ESZSL \cite{eszsl}, ALE \cite{ale}, SAE \cite{sae}, SYNC \cite{sync}, LATEM \cite{latem}, DEM \cite{dem}, DEVISE \cite{devise}, 
  and generative models,
  i.e., SE-ZSL \cite{se-zsl}, f-CLSWGAN \cite{f-clswgan}, CADA-VAE \cite{cada-vae}, JGM-ZSL.
  We observe that generative models perform better than non-generative models in general.
  Synthesizing additional features helps to alleviate the imbalance between seen and unseen classes, which leads to higher accuracy for unseen class samples.
  In contrast, the non-generative methods mostly perform well on seen classes and obtain much lower accuracy on unseen classes, resulting in a low harmonic mean.
  
  Compared with generative methods, our proposed method still achieves state-of-the-art performance on all three datasets, especially in terms of unseen class accuracy.
  W.r.t the harmonic mean, we significantly improve on OCD \cite{ocd} by 11.1\%, 9.2\% and 13.1\% on CUB, SUN and AWA2 respectively.

  \subsection{Ablation Studies}
  
  \subsubsection{Effect of Different Components:}
  
\begin{figure}[!htb]
  \centering
      \includegraphics[width=0.99\linewidth]{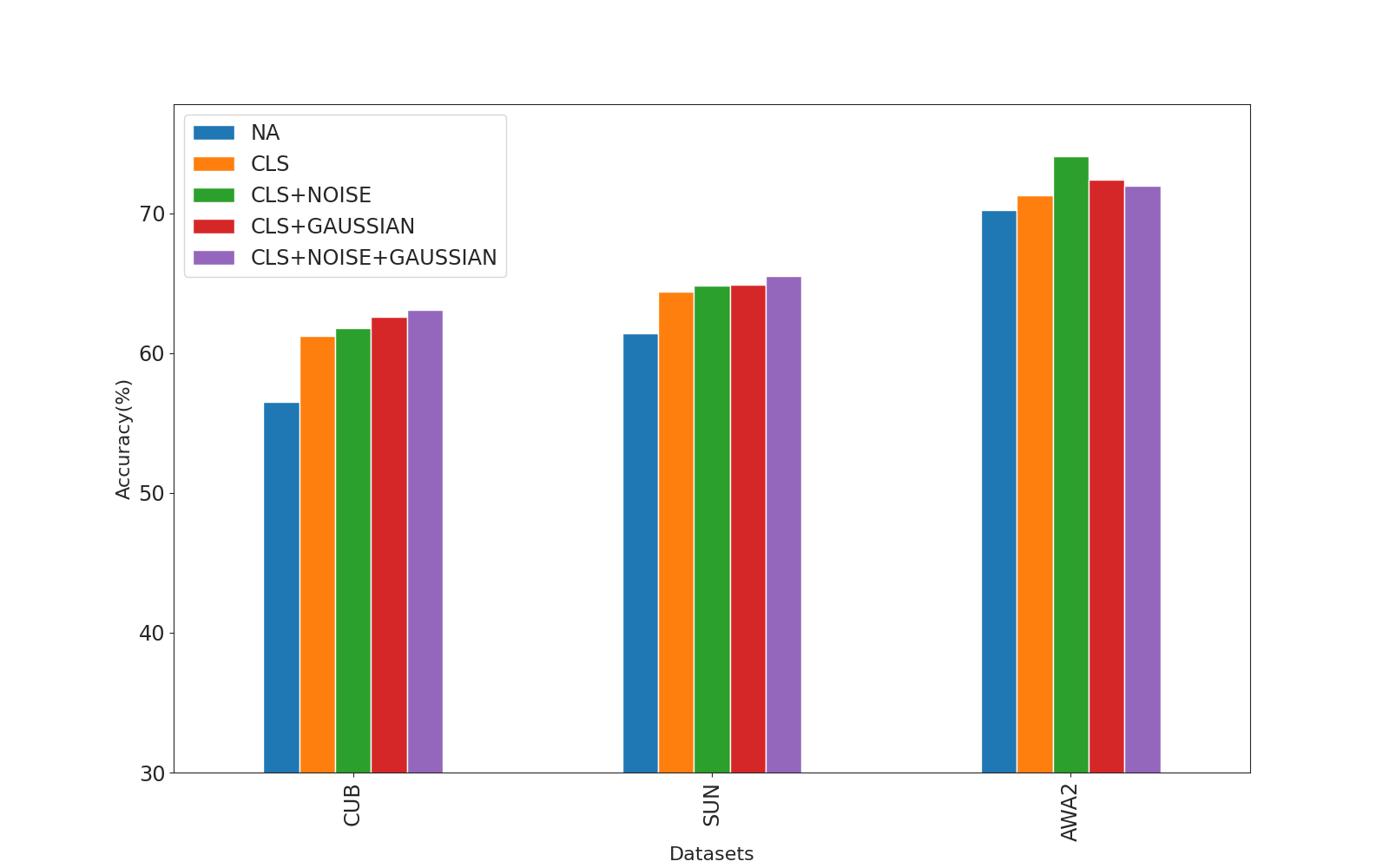}
  \caption{
  Zero-shot learning performance with different components of our method on three datasets. The
reported values are classification accuracy (\%) 
  }\label{fig:ablative}
\end{figure}

 The proposed framework has multiple components for improving the performance of ZSL/GZSL.
 Here we conduct ablation studies to evaluate the effectiveness of each of the component individually.
 Fig. \ref{fig:ablative} summarizes the zero-shot recognition results of different submodels.
 By comparing the results of `NA' and `CLS', we can observe that models improve a lot by projecting the original features to a clusterable space, especially for CUB and SUN. 
 Since CUB and SUN are fine-grained datasets, the original visual features are typically hard to distinguish.
 Therefore, increasing the clusterability of real visual features has a significant effect.
 This can also be seen through the comparisons of `CLS' and `CLS-GAUSSIAN', since fine-tuning with Gaussian similarity loss also leads to better clusterability.
 AWA2 is a coarse-grained dataset, so the original features are already well seperated. Thus the biggest improvement comes from adding noise to obtain a more robust classifer.
 
 \subsubsection{Evaluation of Feature Clusterablility:} Beyond zero-shot learning performance, here we analyze the clusterability of the visual features learned with our proposed method.
 Specifically, we apply one of the most commonly used clustering algorithms, K-Means, on the projected feature space.
 Then we evaluate the clustering performance by computing the mutual information (MI), which measures the agreement of the ground truth class assignments and the K-Means assignments.
 Higher mutual information indicates the clustering algorithm performs better and thus, a more clusterable feature space.
 \begin{table}[t]
    \centering
    \renewcommand{\tabcolsep}{1mm}
    \begin{tabular}{l|cc|cc|cc}
      \hline
       Method & \multicolumn{2}{c|}{CUB} & \multicolumn{2}{c|}{SUN} & \multicolumn{2}{c}{AWA2}\\
        &  MI & Acc & MI & Acc & MI & Acc  \\
      \hline
      \textit{NA}    & 0.67  & 56.6 & 0.56 & 61.4 & 0.83 & 70.2\\
      \textit{Cl}s    & 0.71 & 61.2  & 0.59 & 64.4 & 0.85 & 71.3\\
      \textit{Cls + Gaussian} & \textbf{0.73} & \textbf{62.6}  & \textbf{0.60} & \textbf{64.9} & 0.85 & \textbf{72.4} \\
      \hline
    \end{tabular}  \\ \vspace{6pt}
    \caption{Mutual information (MI) score and classification accuracy on CUB, SUN and AWA2 without our proposed method (\textit{NA}), with the supervision of classification loss (\textit{Cls}) and 
    Gaussian similarity loss (\textit{Gaussian})}\label{tab:eval_score}%
    \vspace{-10pt}
  \end{table}
  
 As seen in table \ref{tab:eval_score}, the proposed mapping network supervised by the classification loss, improves visual feature clusterability  by a large margin, i.e, from 0.67 to 0.71 on CUB, from 0.56 to 0.59 on SUN and from 0.83 to 0.85 on AWA2 in terms of MI score.
 Correspondingly, the zero-shot learning performance also improves from 56.6\% to 61.2\% , from 61.4\% to 64.4\%  and from 70.2\% to 71.3\% respectively.
 Moreover, the fine-tuning step with Gaussian similarity loss further improves MI by 0.02 on CUB and 0.01 on SUN, which brings 1.4\% and 0.5\% zero-shot learning performance improvement.
 
 We can observe from the above experimental results that  feature space clusterability and classification accuracy are strongly correlated:
 more clusterable feature space leads to higher classification accuracy. 
 %
 
  \subsection{Few-shot Learning}
  We also apply our method on few-shot recognition problems to show that more clusterable features can benefit few-shot learning as well.
  A baseline approach \cite{closer-look} to few-shot learning is to train a classifier using the features extracted from a support set.
  The trained classifier is then used to predict labels of query set images.
  We use two fine-grained datasets, CUB and SUN, where we extract  image features by  ResNet101  pretrained on ImageNet.
  We use base set features to train a mapping function supervised by Gaussian similarity loss.
  The mapping function is then applied to a novel set to obtain more clusterable features. 
  We compare our method with the baseline, i.e. using features extracted from the pretrained ResNet directly.
  The results of our experiments as well as comparisons to some baselines \cite{delta-encoder,protonet}
  are summarized in Table \ref{tab:fewshot}.
  We can conclude that increasing features' clusterability can improve the baseline performance by a large margin, especially for the 1-shot setting (16.4\% on CUB dataset and 6.1\% on SUN dataset).
   \begin{table}[t]
    \centering
    \renewcommand{\tabcolsep}{1mm}
    \begin{tabular}{l|cc|cc}
      \hline
       Method & \multicolumn{2}{c|}{CUB} & \multicolumn{2}{c}{SUN} \\
        &  1-shot & 5-shot & 1-shot & 5-shot \\
      \hline
      Ours-Baseline    & 70.2  & 92.6 & 76.5 & 93.1 \\
      ProtoNet   & 71.9 & 92.4 & 74.7 & \textbf{94.8}\\
      $\Delta$-Encoder      & 82.2 & 92.6 & 82.0 & 93.0\\
      Ours-Gaussian & \textbf{86.6} & \textbf{95.8}  & \textbf{82.6} & 93.8 \\
      \hline
    \end{tabular}  \\ \vspace{6pt}
    \caption{1-shot/5-shot 5-way accuracy with ImageNet pretrained features (trained on disjoint cat egories)
    }\label{tab:fewshot}%
    \vspace{-10pt}
  \end{table}

  \subsection{Visualization of Synthesized Samples}
  To further demonstrate how our proposed method boosts zero-shot learning performance, we randomly sample ten unseen categories and visualize the real features and synthesized features using t-SNE\cite{ale}.
  Figure \ref{fig:visualization}  depicts the empirical distributions of the
  true visual features and the synthesized visual features with and without our proposed framework. 
  We observe the original true features contain hard samples close to another class and some of them are overlapping (Figure \ref{fig:vis1}).
  The discriminative classifier trained with synthesized samples typically performs poorly on such a distribution. 
  On the contrary, the projected features are easily separated with larger inter-class distance (Figure \ref{fig:vis2}), which leads to better distribution alignment between real and synthesized features.
  Moreover, by introducing Gaussian noise, the synthesized features exhibit larger intra-class variance (Figure \ref{fig:vis3}), which leads to a more robust classifier to handle the projected test samples.
   
   \section{Conclusion}
   In this work, we have proposed to learn clusterable visual features to address the challenges of Zero-Shot Learning and Generalized Zero-Shot Learning within a CVAE based framework.
   Specifically, we use a mapping function, supervised by softmax classification loss, to project the original features to a new clusterable feature space.
   The projected clusterable visual features are not only more suitable for the generator to reconstruct, but also are more separable for the final classifier to classify correctly.
   To further increase the clusterability of visual features, we utilize Gaussian similarity loss to fine-tune the features first before CVAE reconstruction.
    In addition, we introduce Gaussian noise to enlarge the intra-class variance of the synthesized features to obtain a more robust classifier.
   %
   %
   We evaluate the clusterability of visual features quantitatively and experimentally demonstrate that more clusterable features lead to better ZSL performance. 
   Experimental results on three benchmarks under ZSL, GZSL and FSL settings show our method improves the state-of-the-art by a large margin.
   We are also interested to see how learning clusterable features can be applied to other tasks such as face recognition and person re-identification.
   %

    \bibliography{citation}
  \bibliographystyle{aaai}
  \end{document}